\documentclass[conference]{IEEEtran}
\IEEEoverridecommandlockouts
\usepackage{cite}
\usepackage{amsmath,amssymb,amsfonts}
\usepackage{algorithmic}
\usepackage{graphicx}
\usepackage{textcomp}
\usepackage{xcolor}
\usepackage{enumitem}
\usepackage{subfigure}
\usepackage{nccmath}
\usepackage{amssymb}
\usepackage{mathtools}

\usepackage{booktabs}

\usepackage{hyperref}
\hypersetup{
    colorlinks=true,
    linkcolor=blue,
    filecolor=magenta,      
    urlcolor=cyan,
}

\usepackage[numbers]{natbib}
\begin{document}
\title{The Impact of Federated Learning on Distributed Remote Sensing Archives}
\author{
\IEEEauthorblockN{Anand Umashankar\textsuperscript{*}, Karam Tomotaki-Dawoud\textsuperscript{*}, Nicolai Schneider\textsuperscript{*}}
\IEEEauthorblockA{\textit{Technische Universität Berlin}\\
Berlin, Germany}
\thanks{\textsuperscript{*}Equal contribution, listed in no particular order.}
\thanks{This work was originally completed in 2021 and reflects the state of the literature at that time. It is posted as a historical record and reference baseline.}
}
\maketitle
\begin{abstract}
When it comes to Machine Learning in \textbf{remote sensing}, one of the main obstacles researchers are facing is the large scale of the datasets. 
Just the size of freely available earth observation proposes a challenge for personal computers. For instance, a variety of missions such as Sentinel-1, -2, and -3 gathered a collective of 5 PetaBytes \cite{earthObservationDataManagement}. Given the size of the datasets, they are stored and processed across multiple platforms (often referred to as clients) which implies that decentralized Machine Learning has to be applied. \textit{Federated Learning} is one decentralized learning approach that was first introduced recently\cite{federatedLearningGoogle} by Google and is adopted in their Android ecosystem.
\\
Since its release, the original \textit{Federated Learning} technique has been fine-tuned and further developed. The scope of this project is to apply multiple \textit{Federated Learning} models on remote sensing datasets and understand their implications considering different data splits across clients. The code is available on \href{https://github.com/anandcu3/Federated-Learning-for-Remote-Sensing}{Git}.
\end{abstract}


\section{Introduction}
Remote sensing (RS) datasets are oftentimes too large to be trained on a centralized Machine Learning model. For this matter, the data is split into various partitions and trained separately. One exciting new approach was first introduced by Google researchers in 2017, \textit{Federated Learning} (FL) \cite{federatedLearningGoogle}. 
\\
The idea behind FL is to send the Deep Learning model to the data instead of sending the data to the model. In the case of Google, this method is used to apply Machine Learning on Android devices. The data from each phone is not being sent to a central server. Instead, each device, oftentimes referred to as a \textit{client}, trains a model received from a host/central server based on the client's own data. The trained models from each device are sent back to a central host and averaged.\newline
\\
Accessing the data from different devices is not the root of the issue in our case; however, we consider a bigger dataset and split it into a variety of partitions to apply FL. The approach might solve the issue of training big datasets; nevertheless, it also comes along with two main challenges:
\begin{itemize}
  \item The first obstacle is the extensive communication between clients and the host for model averaging, which can highly drain the training process.
  \item The second hurdle arises through the client data distribution. Considering a remote sensing dataset with images from all over the world, there are certain classes like 'desert' that can only be found in a few regions of the world. In case the data is distributed by country, most clients wouldn't have access to such classes (as 'desert'). This characteristic is also called non-IID (non-independent and identically distributed) data partition. \cite{non_iid_main_paper}
\end{itemize}

Over the past years, a variety of FL approaches have been developed to tackle these issues. For instance, \textit{FedAvg} \cite{fedAvg} decreases the client-server communication by only training a randomly chosen fraction of clients during each epoch. Another approach is \textit{FedProx} \cite{fedProx}, which addresses the hurdle of non-IIDness by adding a \textit{proximal term} to consider the degree of IIDness of each client during training. The goal of this project is to apply these FL approaches using different data partitions to both understand the impact of Federated Learning on non-IIDness and how different data distributions can affect the results.

\subsection{Goals and Challenges}
Federated Learning is still a new topic both in the world of academia and the industry. When applied correctly it can solve many issues, but it also proposes new challenges. We intend to implement three different Federated Learning models (\textit{Bulk Synchronous Parallel (BSP)} \cite{BSP_orig_paper}, \textit{Federated Averaging (FedAvg)}, and \textit{Federated Proximal(FedProx}) on a RS dataset to understand their impact in comparison to an ordinarily used centralized approach. All implementations will be tested with the Deep Learning models \textit{ResNet34} \cite{DBLP:journals/corr/LimSKNL17}, \textit{AlexNet} \cite{KriSut12Imagenet} and \textit{LeNet} \cite{lecun-gradientbased-learning-applied-1998}. 

We evaluate if these federated learning algorithms are effective on remote sensing datasets. We intend to make comparisons among different deep learning models when using federated learning. Lastly, we would like to modify different hyperparameters and other experiment settings to evaluate the extent of the effects that these have on the outcome. The main criteria for these comparisons are the accuracy of the output models as well as communication costs and runtime. Based on these comparisons we empirically conclude the optimal federated algorithm, deep learning model, and hyperparameter choices that can be used for future RS applications.

Classical RS datasets tend to be very large, making the computational process much more difficult. Nevertheless, this issue goes beyond the scope of our project, therefore we chose UC Merced Landuse \cite{UCMDATASET}, a multilabel RS dataset containing 2100 images and 18 classes.
\\
We expect to gain similar results to the current literature and to find the optimal parameters for each FL model.

\section{Background and Related Work}
\label{section:Background}

The first section provides information about the basic implementation of Federated Learning, the chosen FL algorithms, and the applied Deep Learning models for our experimental evaluation. We then discuss current findings and approaches in Federated Learning.

\subsection{Federated Learning}
\begin{figure}[h]
    \centering
    \includegraphics[width=\linewidth]{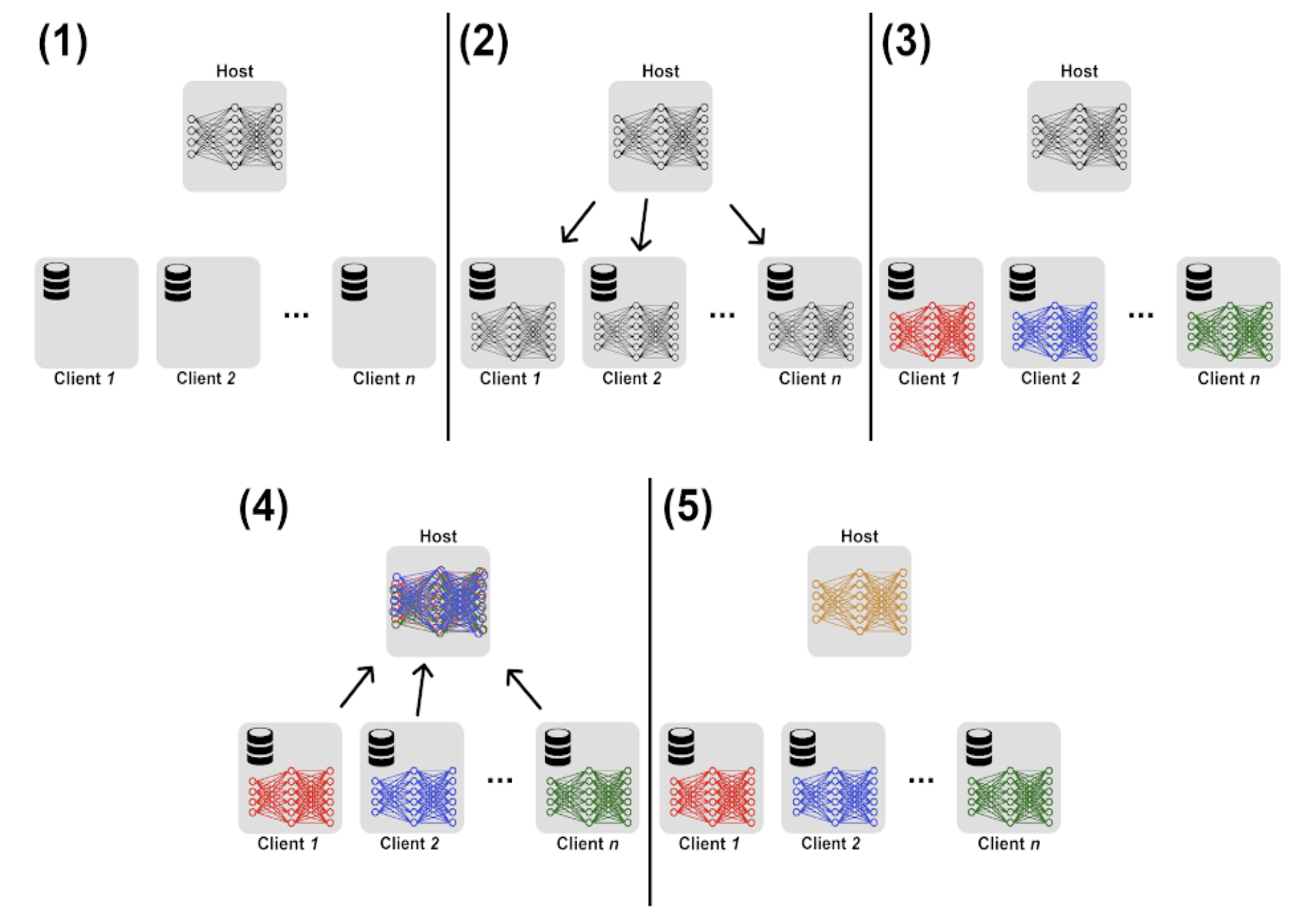}
    \caption{Basic steps of Federated Learning}
    \label{fig:basic_fl}
\end{figure}

The main idea of \textit{Federated Learning} is to reverse the common procedure of Machine Learning: instead of sending the data to the model, the model is sent to the data. In an FL scenario, we have two parties: the \textit{host} and the \textit{clients}. The host contains the Deep Learning model, which will later be trained, while each client holds a fraction of the dataset. The main steps are depicted in Fig. \ref{fig:basic_fl}. In step 1, the \textit{host} initializes a Machine Learning model and sends it to each \textit{client} (step 2). Next, each \textit{client} trains the received model based on its data (step 3) and sends the trained model back to the \textit{host} (step 4). The \textit{host} then collects all models and averages them (step 5). It should be noted that the training of each client takes place in parallel.

\subsubsection{Federated Averaging}
\begin{figure}[h]
    \centering
    \includegraphics[width=\linewidth]{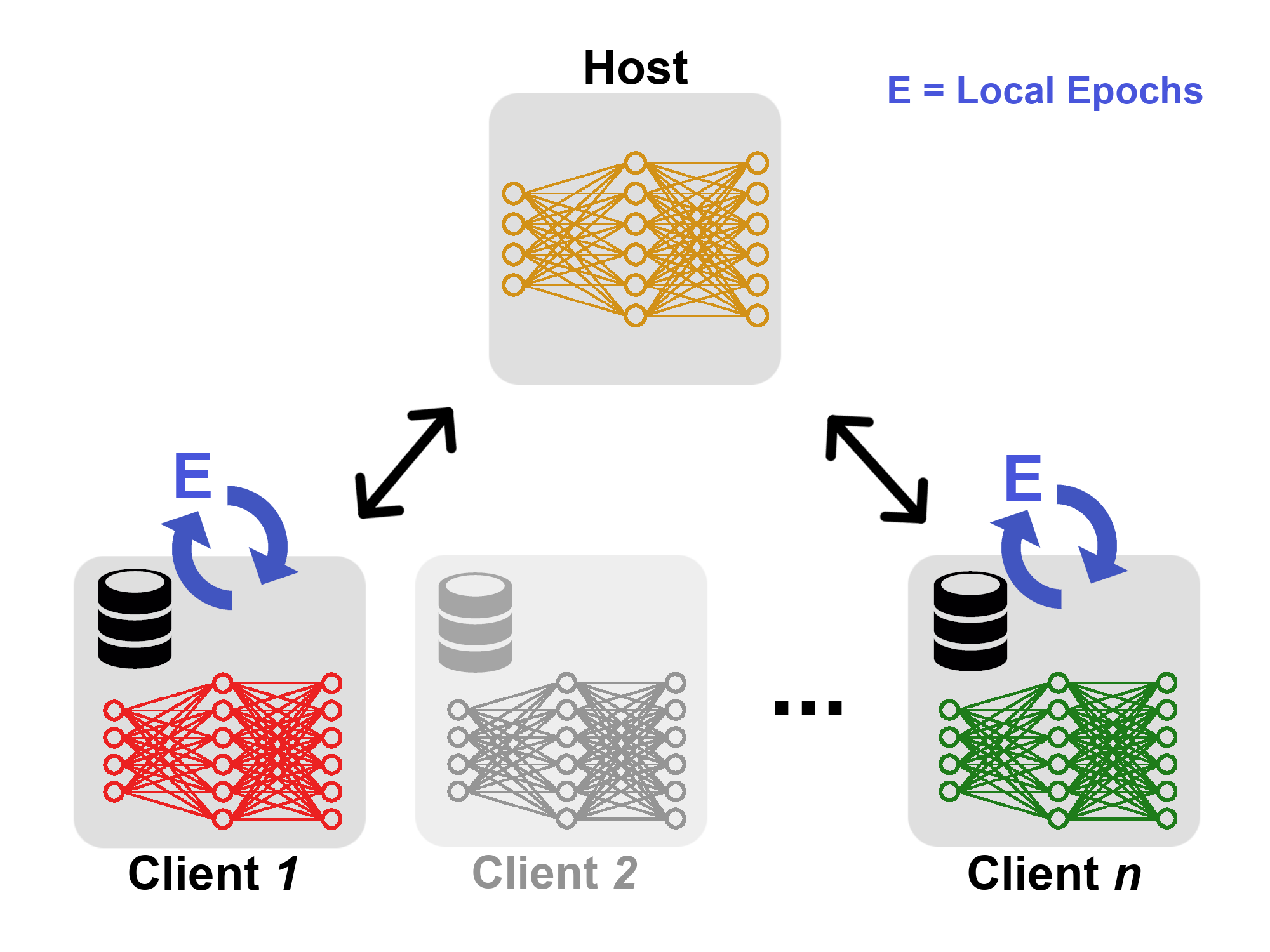}
    \caption{Basic steps of Federated Averaging}
    \label{fig:fedavg}
\end{figure}

The basic \textit{Federated Learning} model presents one major issue, which is the enormous communication between the \textit{clients} and the \textit{host}, and the high computation. One of the most common Federated Learning algorithms, which tries to tackle these issues, introduced in \cite{fedAvg}, is \textit{FedAvg}.
\\
Let $K$ be the set of clients, for each training round \textit{FedAvg} only sends the model to a random fraction with a fixed size $C\subseteq{K}$ of clients. For instance, for the experimental evaluation of \cite{fedAvg}, only 10\% of clients were trained each round. Furthermore, the communication is reduced by running multiple \textit{local epochs} $E$ as depicted in Fig. \ref{fig:fedavg}. The authors used up to 5 local epochs for their experiments. Finally, each client's local dataset can be split into batches by applying the parameter $B$, where $B=\infty$ specifies that the whole local dataset is used as a batch. Once all clients $k\in{C}$, with their respective data partition $n_k$, sent their trained weights ${w_k}^t$ back to the host, the new averaged model $w_{avg}^t$ is computed with
\begin{equation}
    w_{avg}^t=\sum_{k=1}^{|K|}\dfrac{n_{k}}{n}w_{k}^t
\end{equation}
where $t$ indicates the training round and $n$ the length of the whole dataset.

\subsubsection{FedProx}
\textit{FedProx} \cite{fedProx} is an extension to \textit{FedAvg} that has modifications to tackle non-identical distributions in data and also accounts for system heterogeneity. \textit{FedProx} provides more reliable convergence when compared to \textit{FedAvg}. On average, a 22\% accuracy improvement is shown across highly heterogeneous settings. Their work is mainly based on adding a "proximal" term to a standard local loss function. The objective is the usual loss function, summed with a penalty when the local model deviates too much from the global model. This addresses the issues of data heterogeneity and allows for safely incorporating variable amounts of local work resulting from system heterogeneity.

\subsubsection{Bulk Synchronous Parallel}
\textit{Bulk Synchronous Parallel} (\textit{BSP}) \cite{BSP_orig_paper} is an older approach that misses the key FL element of averaging the models. In terms of \textit{FedAvg} the parameters are set the following way: 
\begin{itemize}
    \item $C=1$, therefore all clients are used in each round
    \item $E=1$, such that each client runs 1 local epoch
\end{itemize}
Instead of passing the model to each client and averaging the trained models, \textit{BSP} passes the model from one client to the other. Once a client is done with training, it sends the model to the next client. A round is complete once the model has been passed to each client. A more communication-heavy version of BSP will pass the model between clients after training on a single training batch. This communication-costly approach is more robust to the non-identical distributions in data since it takes more small update steps towards convergence instead of large updates that might skew the model in one direction or the other.
\subsection{Deep Learning Models}

\textit{LeNet}
 is one of the earlier Machine Learning approaches and was first proposed in 1990. The original architecture of \textit{LeNet-5} consisted of two convolutional layers, two sub-sampling layers, two fully connected layers, and an output layer with Gaussian connection \cite{lecun-gradientbased-learning-applied-1998}. To adapt to the image size of 256x256, we adjusted the kernel size for all convolutional layers to 5x5.

\textit{AlexNet} was first introduced by Alex Krizhevesky in 2012 and was considered a State-of-the-Art Deep Learning model for visual recognition and classification at the time. The architecture consists of a total of 8 layers: five convolutional layers, two fully connected layers with dropout, and a softmax layer.

\begin{figure}[h]
\centering
    \includegraphics[width=\linewidth]{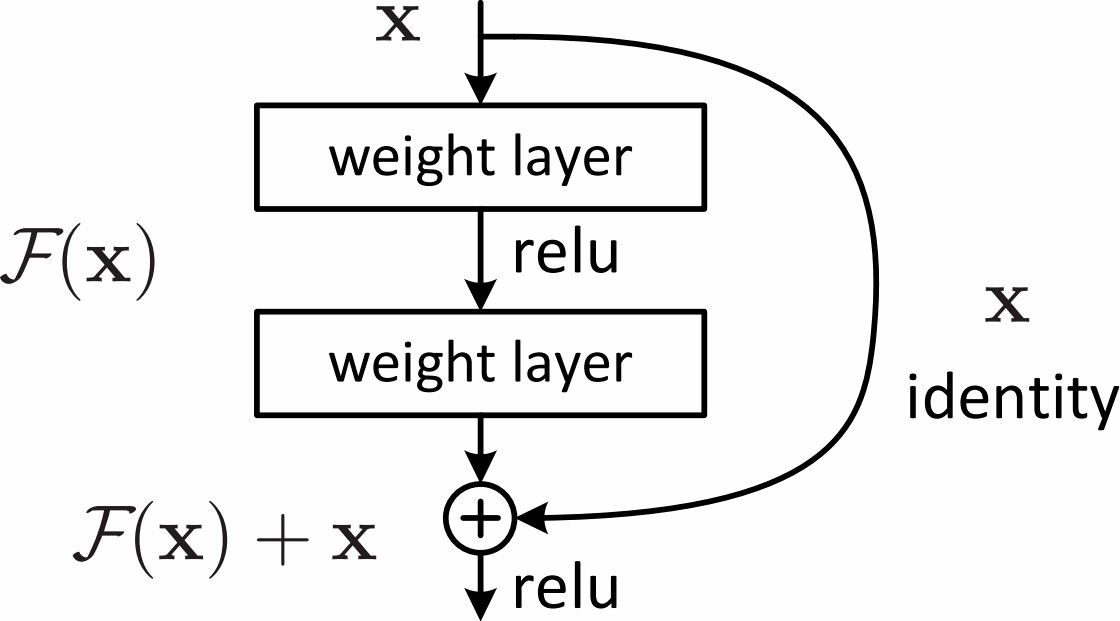}
    \caption{Residual block used by ResNet architecture}
    \label{fig:Residual_block}
\end{figure}

\textit{ResNet} is one of the most popular approaches in Image classification and has been published in 2015 by Kaiming He. The main architecture consists of convolutional layers with a 3x3 filter and concludes with an average pooling layer and a 1000-way fully-connected layer with softmax. Additionally, ResNet stacks building block (shown in Fig \ref{fig:Residual_block}), using the so-called shortcuts to skip the input over the next two layer which makes the CNN \textit{residual} \cite{DBLP:journals/corr/LimSKNL17}. The shortcuts can only be used when the input and the output have the same dimensions, and they help solve vanishing gradients problem, which is one of the main problem in training deeper and deeper Neural Networks. 

\subsection{Related Work}
\cite{non_iid_main_paper} shows that training over skewed label partitions is a challenging problem to solve, especially for decentralized learning, as all the algorithms in their study suffer major accuracy loss. Secondly, DNNs with batch normalization were found to be vulnerable in the Non-IID setting. They also prove that the difficulty level of this problem varies greatly with the degree of skew. They use three decentralised training algorithms, which are Gaia \cite{Gaia10.5555/3154630.3154682}, Federated Averaging, and Deep Gradient Compression \cite{lin2020deep}. 

\subsection{Other FL Algorithms}
\textit{Gaia} \cite{Gaia10.5555/3154630.3154682} accumulates updates to model weights and updates it to other data partitions when its relative magnitude exceeds a defined threshold, which means that the insignificant communication between data centers is reduced while still retaining correctness of machine learning approaches. They observed speedup of almost 1.8x to 53.5x over leading distributed ML frameworks, and it is 0.94x - 1.4x when using the same ML approaches on nodes connected on a local area network. 

\textit{Deep Gradient Compression} \cite{lin2020deep} communicates only a pre-specified amount of gradients each training step to reduce communication costs. This is also called as gradient clipping and is done on the local nodes. They also use other approaches like momentum correction, momentum factor masking, and warm-up training. In their experiments they achieve a compression ratio of 270x to 600x without losing accuracy. 

\textit{SCAFFOLD} \cite{karimireddy2020scaffold} uses a variance reduction technique to correct the drift off in local clients in its local updates. SCAFFOLD requires significantly lower communication rounds when compared to \textit{FedAvg} and also performs well irrespective of data heterogeneity or client sampling. SCAFFOLD can also take advantage of similarity in different clients' data, thus resulting in even faster convergence in those cases. Their experiments prove that they are always at least as fast as normal SGD and can be much faster depending on the data similarity between clients.

\textit{FedBoost} \cite{pmlr-v119-hamer20a} provide ensemble algorithms that are made optimised to have low communication for federated learning. In their work the per-round communication cost is independent of the size of the ensemble. Unlike other previously discussed works \cite{lin2020deep} \cite{fedAvg}, their approach reduces the communication between both server-to-client and client-to-server communication. 

\textit{FetchSGD} \cite{rothchild2020fetchsgd} compresses model updates using Count Sketch. This enables the solution to take advantage of the combinability of the sketches to combine model updates from many nodes into one update. The Count Sketch is linear in nature, and hence, momentum and error accumulation can be performed inside the sketch. This helps to move the momentum and error accumulation from clients to the central aggregator, thus solving the problems associated with client participation and also achieving high compression rates and good convergence.

\section{Methodology}

\subsection{Dataset \& data augmentation}
\begin{figure}[h]
\centering
    \includegraphics[width=.92\linewidth]{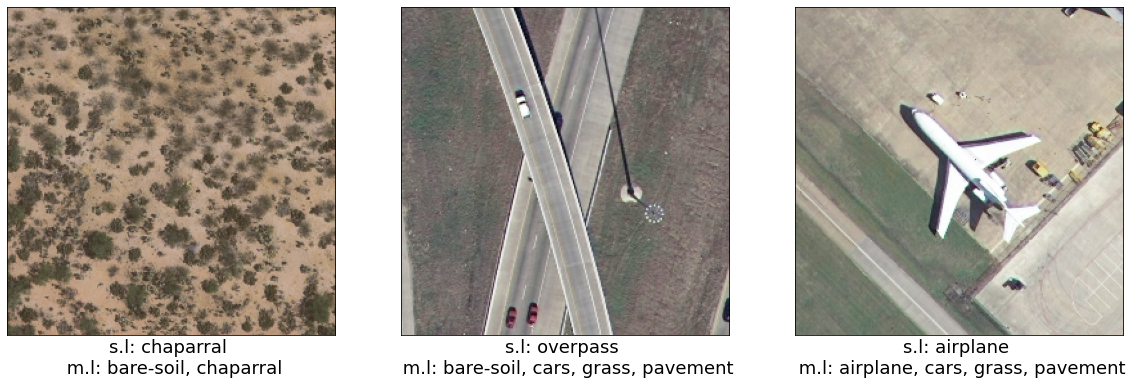}
    \includegraphics[width=\linewidth]{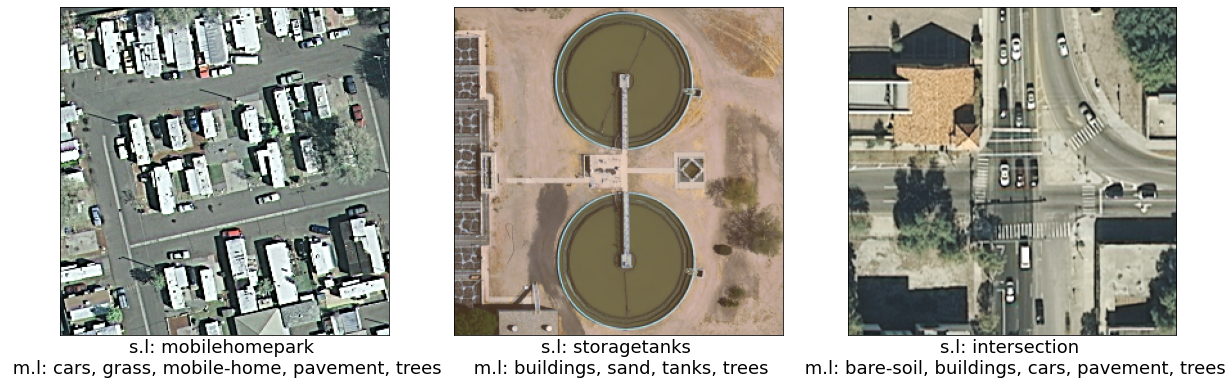}

    \caption{Some UC Merced Land Use Dataset examples, showing both the original single label (s.l) as well as the multilabel(m.l).}
    \label{fig:UCMerced_ex}
\end{figure}

Our Dataset of choice for this experiment is \href{http://weegee.vision.ucmerced.edu/datasets/landuse.html}{UC Merced Land Use Dataset} \cite{UCMDATASET}, but instead of using the provided single label, we opt for using the \href{https://bigearth.eu/datasets.html}{multilabel} \cite{MultilabelUCM}, because multilabel are usually more realistic and challenging for a Remote Sensing classification case study (examples shown in Fig \ref{fig:UCMerced_ex}).\newline
The dataset contains 2100 images, which is a small number for training, especially when using a large number of clients. Therefore, before training, we used data augmentation to double the dataset in size to 4200. We apply one of four common corruption methods on each image once; "\textbf{Impulse noise} is a color analogue of salt-and-pepper noise and can be
caused by bit errors...\textbf{Motion blur} appears when a camera is moving quickly...\textbf{Snow} is a visually obstructive
form of precipitation...\textbf{Pixelation} occurs when up-sampling a low-resolution image." \cite{hendrycks2019benchmarking}.
Furthermore, during training, a random horizontal flipping was also applied.

\begin{figure}[h]
\centering
    \includegraphics[width=.3\linewidth]{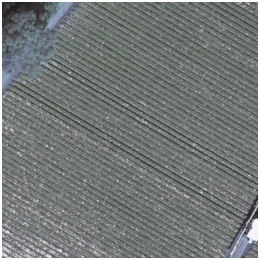}
    \includegraphics[width=.3\linewidth]{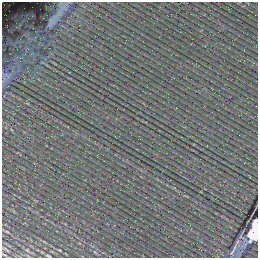}
    \includegraphics[width=.3\linewidth]{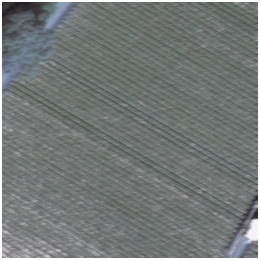}
    \includegraphics[width=.3\linewidth]{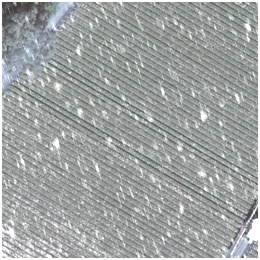}
    \includegraphics[width=.3\linewidth]{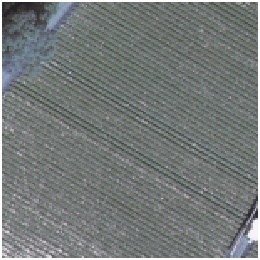}

    \caption{Example of the different augmentation methods on the same image, in the first row from left to right: original image, impulse noise, motion blur. In the second row: snow, and pixelation.}
    \label{fig:UCMerced_augment}
\end{figure}

\subsection{Main aspects of our experiment}
\label{section:exps}
\begin{figure}[h]
\centering
    \includegraphics[width=0.9\linewidth]{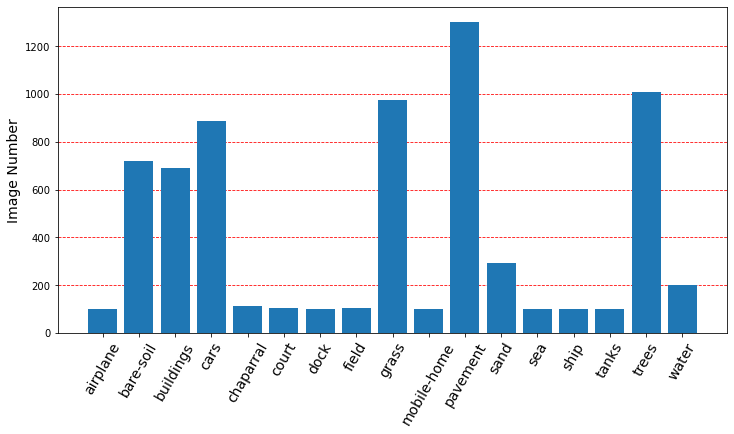}
    \caption{UC Merced Land Use Dataset multilabel distribution: the total number of label occurrences in the 2100 images of the  Dataset. We define "common labels" as labels that are present in more than 10 \(\%\) (~210 data points), whereas "less common labels" are in less than 10 \(\%\). }
    \label{fig:UCMerced_dist}
\end{figure}

\begin{figure}[h]
\centering
    \includegraphics[width=0.9\linewidth]{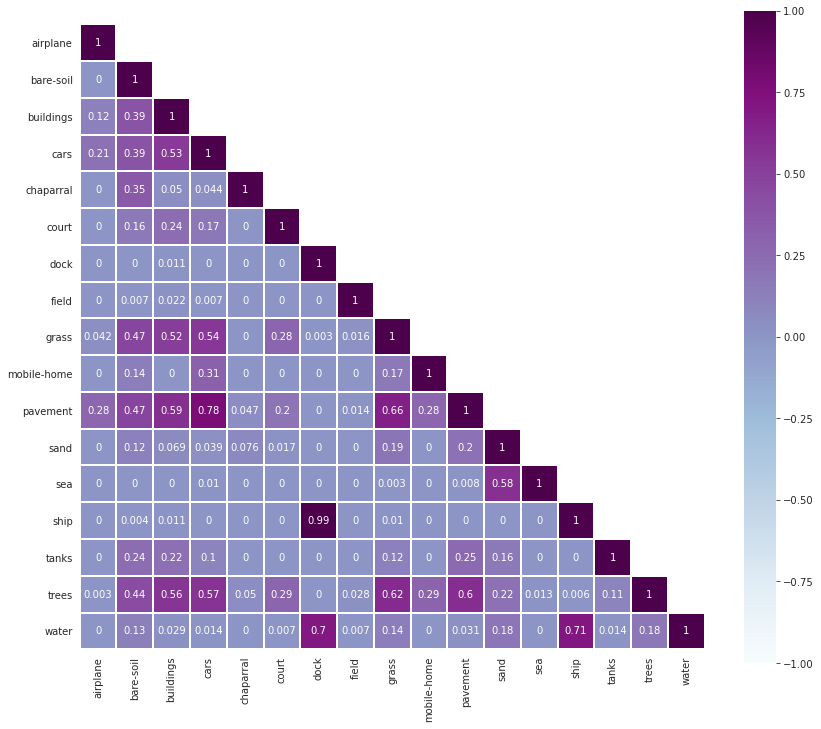}
    \caption{UC Merced Land Use Dataset multilabel cosinus similarity matrix; shows that "less common" labels are for the most part decorrelated, whereas "common labels" are much more correlated. The darker purple a matrix field gets (closer to 1), the more correlation (co-occurrences) between two labels there are, whereas the bluer (at zero), the 2 labels never exist in the same image.}
    \label{fig:UCMerced_similarity}
\end{figure}

Similar to \cite{non_iid_main_paper}, our study focuses on the following criteria: 
\begin{enumerate}[label=(\roman*)]
    \item The ML models: for this we compare the influence of FL on the validation accuracy for different neural networks. AlexNet \cite{KriSut12Imagenet}, LeNet \cite{lecun-gradientbased-learning-applied-1998}, and ResNet34 \cite{DBLP:journals/corr/LimSKNL17}. Training parameters were set to learning rate = 0.001 and Momentum = 0.9 for all the models.
    
    \item Federated Learning algorithms: as described in \ref{section:Background}, we compare FedAvg and FedProx against each other as well as against BSP. For FedAvg, we used the following hyper-parameters: \(C_{fraction} \in \{0.5, 0.75, 1\}\) (meaning in each round the model is sent to half, three-quarter, and all clients for training which effectively reduces the amount of data used for each round of training), with local epoch number on each client \(E_{local}  = 5\).
    
    \item Degree of label skew-ness of the datasets partitions: the idea here is that each client has a monopoly of some percentage over a certain label in the dataset, whereas the rest of the dataset is uniformly distributed over all the clients.
     \newline 
    But there is an inherent problem in artificially skewing multilabel datasets over a certain number of clients. As seen in the label distribution in Fig. \ref{fig:UCMerced_dist}, the dataset has 2 clear types of label dominance, so there is two cases for skewing: 
    \begin{itemize}
        \item common labels: there is only 7 labels that are present in more than \(10\% \) (6 of them are in more that \(25\% \)) of the images, which mean if we distributed the dataset over 4 clients for example there is only a certain degree of skew-ness possible before the label overlaps and the skewness loses its meaning because of the high correlation\footnote{As shown in Fig \ref{fig:UCMerced_similarity}, using the Cosinus similarity measurement clearly shows that the common labels co-occur in the same image much more than less common labels.} between these labels. For our tests, when splitting over those dominant labels, we use 4 clients and skewness \(\in \{0, 20, 40\%\}\).
        \item less common labels: 9 labels are present in roughly \(5\% \) and one label is present in around \(10\% \) of the dataset, and they are highly uncorrelated, which means we can freely skew the monopoly of the clients to a higher percentage, and we can use more clients in this case. In our tests, we used mainly 8 clients with skewness \(\in \{40, 60, 80\%\}\). We also tested increasing the number of clients to \(\{10, 25, 50\}\), with skewness of 40\%, which means the first 9 clients will have the 40\% monopoly over the small 9 labels and the rest of the dataset is uniformly distributed over all the clients. We can see here that for this dataset, as we increase the number of clients being used, the data distribution becomes more IID in nature.
    \end{itemize}
    Furthermore, we don't consider using a mix of common and less common labels for splitting over the clients, since it will cause an imbalanced distribution of data among clients, which is a different kind of FL problem that we are not tackling in this work.

 \item Furthermore, we test the influence of the training batch size on such a setup, with batch sizes \(\in \{1, 4, 8, 16, 32, 64, 128, 256\}\)
\end{enumerate}
 
\subsection{Experiments}
To evaluate all aspects mentioned in \ref{section:exps}, we divide our experiments into 8 sections. The parameters for all our experiments are noted in Table \ref{table:experiments}\footnote{In the table \ref{table:experiments}, the flag called \textit{Small Skew} refers to skewing over the less common label classes.}. Each training runs for 100 Rounds and $E_{local}=5$ for \textit{FedAvg} and \textit{FedProx}. 
\begin{enumerate}[label=(\roman*)]
    \item The first experimental section analyzes a centralized Machine Learning training and BSP using \textit{LeNet}, \textit{ResNet} and \textit{AlexNet} to get a picture of their impact without using Federated Learning.
    
    \item In the following section, we compare the impact of different $C_{fraction}$. We use \textit{FedAvg} and run each Deep Learning model with \(C_{fraction} \in \{0.5, 0.75, 1\}\) and 8 clients.
    
    \item The next part of the experiment considers each Deep Learning model on \textit{FedAvg} using 8 clients and $C_{fraction}=0.75$. This examination increases the \textit{Skewness} in comparison to other experiment sections to 60\% and 80\%.
    
    
    \item This section focuses on a smaller skew percentage with \textit{Skewness} set to 40\%. 20\% and 0\%. We use each of the three Deep Learning models and apply them on \textit{FedAvg} and \textit{BSP} with 4 clients.
    
    \item We compare the impact of different client numbers in this experimental section. For each model we run a training with client numbers $n\in{10,25,50}$ on \textit{FedAvg} with $C_{fraction}=0.5$.
    
    \item We repeat the experiment from (v) using \textit{FedProx}.
    
    \item Finally, we measure the weight of different batch sizes $bs$ with $bs\in{1,4,8,16,32,64,128,256}$ on \textit{FedAvg} with 4 clients using \textit{LeNet}.
\end{enumerate}

\subsection{Evaluation metrics}
Simple Accuracy is defined as:
\begin{equation}
\label{eq:simpleAccuracy}
\text { Accuracy }=\frac{1}{m} \sum_{i=1}^{m} \frac{\left|Y_{i} \cap Z_{i}\right|}{\left|Y_{i} \cup Z_{i}\right|}
\end{equation}
where \(Z_{i}\) denotes the model prediction for the data point \(x_{i}\), \(Y_{i}\) denotes the true label of \(x_{i}\), and  \(i \in \{1,..., m\} \). This measure, however, can be misleading in measuring the quality of the learned model for multilabel applications (also depends on the nature of the dataset). For example, in the UC Merced LandUse multilabel dataset, using this race metric, one can achieve \(80\%\) by predicting the single label \textit{pavement} for all the images. Hence, this was eventually dropped from our final evaluation metric.

Other metrics such as Classification Accuracy \cite{inproceedings}, defined as:   

\begin{equation}
\label{eq:CAccuracy}
\text { ClassificationAccuracy }=\frac{1}{m} \sum_{i=1}^{m} \delta\left(Z_{i},Y_{i}\right)
\end{equation}
where \(\delta = 1\) only if the prediction match the true label for all the labels otherwise  \(\delta = 0\), can be too rigid of a metric.
\newline 
In general the evaluation of methods that learn from multi-label data requires different measures than those used in the case of single-label data \cite{MaimonOdedZ2005Dmak}, those evaluation measurements can be divided into example-based, label-based, and ranking-based.\cite{MaimonOdedZ2005Dmak}
\newline 
For our experiment, we use one of the label-based measurements, that is, the harmonic mean between precision and recall, also known as F1 score \ref{eq:f1score}, which can also be used for evaluating a single-label classifier.  

\begin{equation}
\label{eq:f1score}
F_{1}= \frac{1}{m} \sum_{i=1}^{m} \frac{2\left|Y_{i} \cap Z_{i}\right|}{\left|Z_{i}\right|+\left|Y_{i}\right|} 
\end{equation} 

\begin{figure*}[h]
\centering
    \includegraphics[width=\linewidth]{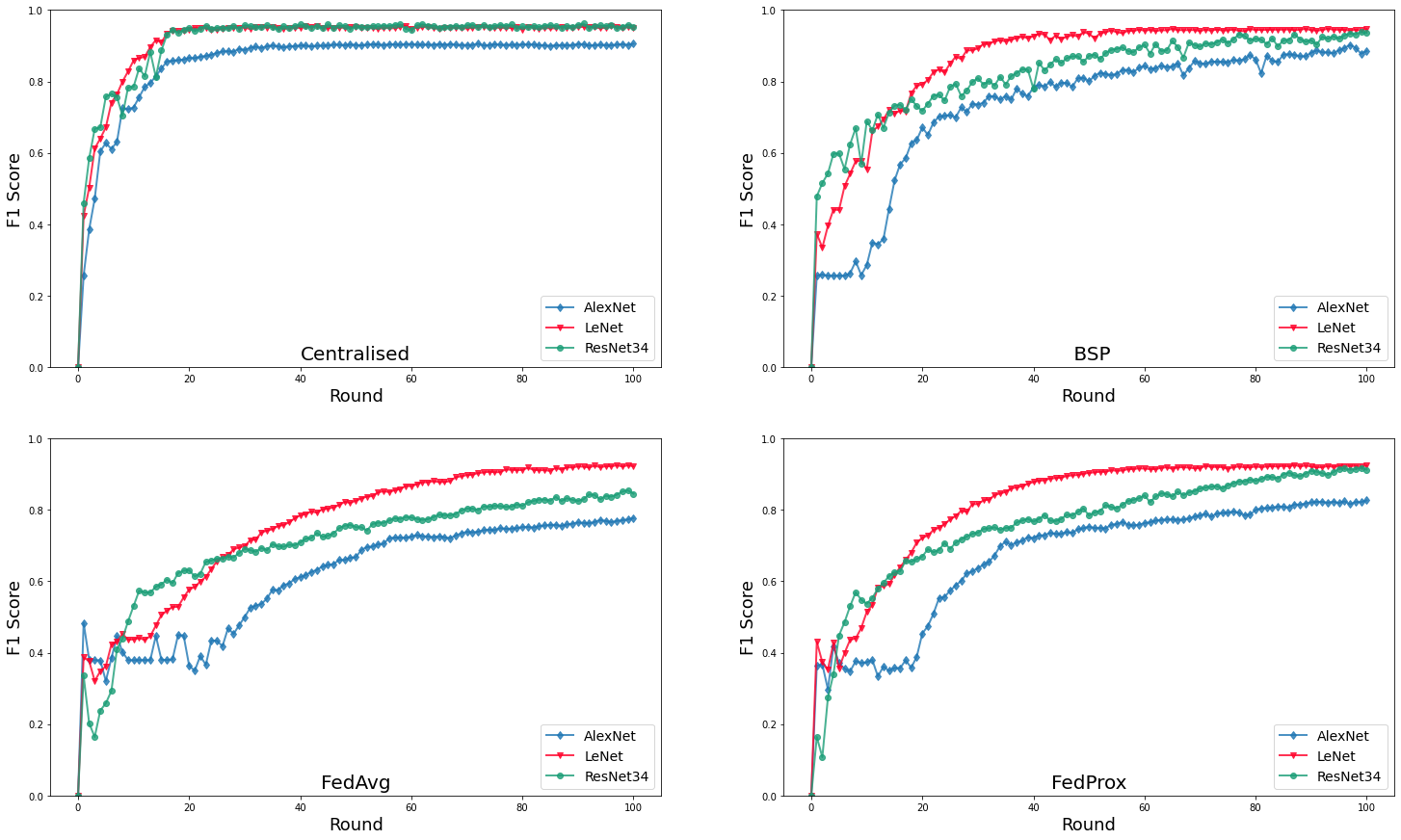}

    \caption{Comparing centralized vs BSP and federated Learning. For BSP, FedAvg, and FedProx 8 clients \(40\%\), skewness on the less common labels was used. The \(c_{fraction}\) for FedAvg and FedProx is set to 0.75 here.}
    \label{fig:CentvFed}
\end{figure*}

\subsection{Implementation Details}
The implementation is done completely in Python. It can be accessed \href{https://github.com/anandcu3/Federated-Learning-for-Remote-Sensing}{in our GitHub repository}. We use PyTorch \cite{CollobertTorch7:Learning} as the preferred choice of Deep Learning Framework. To simulate the different clients for federated learning, we initially considered using PySyft \cite{openmined}, but ran into many issues because of the nascent nature of the library. It was incompatible with the multilabel dataloader on PyTorch and also did not support custom data splitting for the data on different clients. These challenges proved to be too big, and we then decided to use PyTorch directly to simulate the clients and concentrate more on the implementation of the federated algorithms. We use Pandas, matplotlib, numpy packages for the visualisation of the data and generate line plots, bar graphs, etc.

\subsection{Experimental Setup}
The experiments were conducted on a workstation with Intel(R) Xeon(R) W-2133 12 core CPU @ 3.60GHz with 64GB RAM. It was equipped with a NVIDIA GeForce RTX 2080 12GB GPU. It used Ubuntu 18.04 with CUDA and cuDNN for GPU acceleration.

\section{Experimental Results and Discussion}

This paper aims to analyze the effect of decentralized learning on different deep learning models for multi-label remote sensing data. In this section we present our results, showing federated model quality differences compared to centralized learning for the three Deep Learning models (\textit{AlexNet}, \textit{ResNet} and \textit{LeNet}). Then we look at the influence of hyperparameters and other settings such as batch size, \(c_{fraction}\), number of clients, and the degree of skewness for Federated Averaging. 

\subsection{Overall Training Results}

We present the overall training results with two main criteria in mind. The F1-score and convergence time.

\subsubsection{Centralized v. Federated}As shown in Fig \ref{fig:CentvFed}, centralized learning converges better and quicker than all the federated learning algorithms in terms of F1 quality score. But it is important to keep in mind that for centralized learning, skewness is not considered at all, and a direct comparison to federated learning is unfair. Centralized training results however must be seen as benchmark results and not used for direct comparison.

\subsubsection{Comparison of Federated Algorithms}

BSP started to converge fast in the first 20 training rounds, and it slowed down thereafter, but steadily approached the upper bound of the centralized learning. This pattern is the same for all the three Deep Learning models.

\begin{figure}[h]
\centering
    \includegraphics[width=\linewidth]{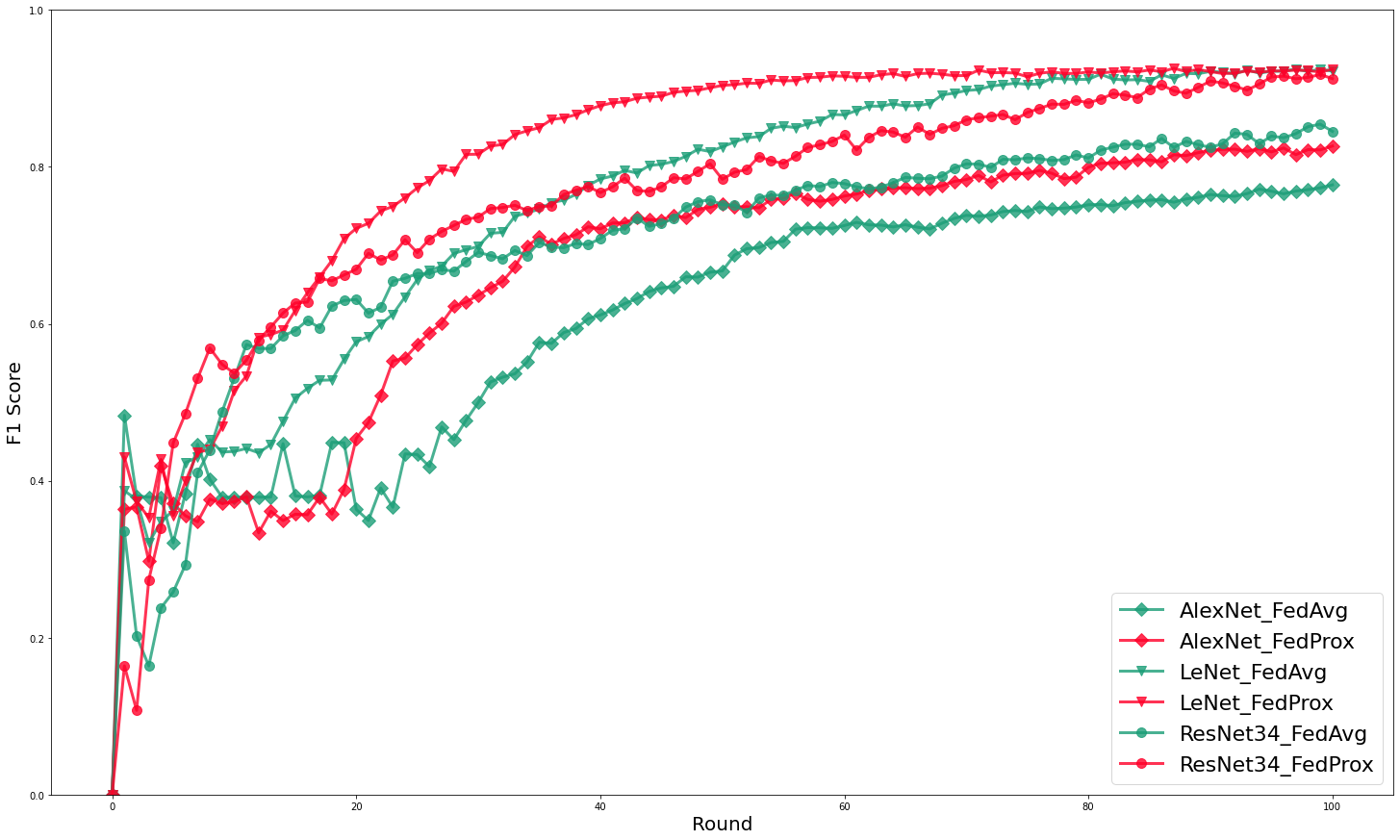}

    \caption{FedAvg vs FedProx for different Deep Learning Models. 8 clients with a \(c_{fraction}\) of 0.75, as well as \(40\%\) skewness on the less common labels were used here.}
    \label{fig:AvgvProx}
\end{figure}

Both FedAvg and FedProx are much slower in converging (than BSP) for most deep learning models. They also fail to reach the upper bound of centralized learning results, sometimes even after training 100 rounds. In direct comparison between FedAvg and FedProx, seen better in Fig \ref{fig:AvgvProx}, it is clear that FedProx (in pink plots) has the upper hand in both quality and convergence speed. That is because of FedProx's loss function being able to keep the clients in check and prevent diverging from the central average model, thus being capable of handling different data distributions and skewness better than FedAvg.   

\subsection{Comparison of Deep Learning Models}
We compare the three deep learning models, i.e., LeNet, ResNet, and AlexNet, based on the results presented in Fig \ref{fig:CentvFed}. For all the federated and centralized training experiments, AlexNet performs the worst in terms of F1 score. This could be because AlexNet is a very large model and requires large datasets to be trained correctly. Given that our dataset size is quite small, even with augmentation, AlexNet needs the dataset to be much larger. Out of the other two models, LeNet generally converges very quickly, compared to ResNet. This could again be since LeNet is a small model and hence is quite suitable for our application. Finally, ResNet manages to converge to the same level as LeNet for BSP. In FedAvg, however, ResNet lags behind LeNet. This is again fixed using FedProx, which has better convergence for ResNet.

\subsection{Drill-down Experiments for Federated Averaging}
In this section, we take a deeper look into results using \textit{Federated Averaging} and varying different hyperparameters and other settings to analyze the effect that it has on the convergence of the deep learning models. For each of these sets of experiments, we vary one of the parameter, while keeping all the other values and settings the same. 

\subsubsection{Client Fraction}
\begin{figure}[!ht]
\centering
    \includegraphics[width=\linewidth]{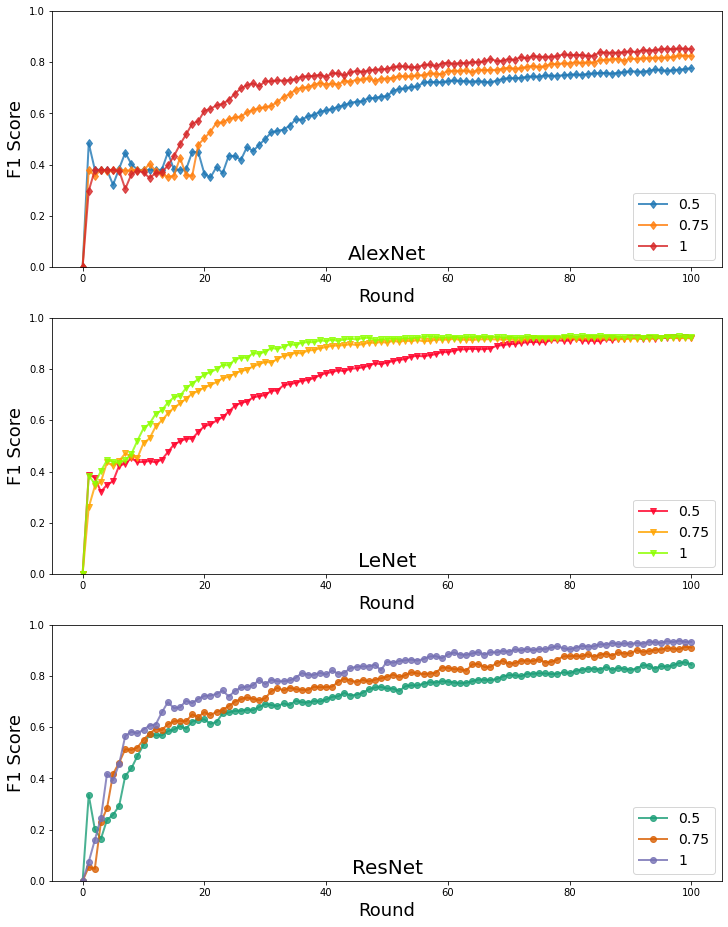}
    \caption{Effect of varying \(C_{fraction} \in \{0.5,0.75,1\}\). 8 clients, and \(40\%\) skewness on the less common labels were used.}
    \label{fig:c_frac}
\end{figure}

We vary the Client Fraction (\(c_{fraction}\)) between 0.5, 0.75, and 1. We use 8 clients for these experiments and consider the less common labels to maintain equal data distribution among the clients. Looking at Fig. \ref{fig:c_frac}, the training using \(c_{fraction} = 1\) converges the best, which was expected since it uses all the clients and effectively all the training data during each round. Predictably, the convergence drops for \(c_{fraction} = 0.75\) for ResNet and AlexNet. Setting \(c_{fraction} = 0.5\) for these models further deteriorates the F1-score. It is remarkable that LeNet is still able to achieve optimum results with \(c_{fraction} = 0.5\), even though it takes longer to converge. Even though the accuracy drops occur with lowering the client fraction, it should be taken into consideration that this also reduces communication costs, which can help to manage bandwidth costs. This will be further discussed in Section \ref{section:commcost}. 

\subsubsection{Number of clients}
\begin{figure}[h]
\centering
    \includegraphics[width=\linewidth]{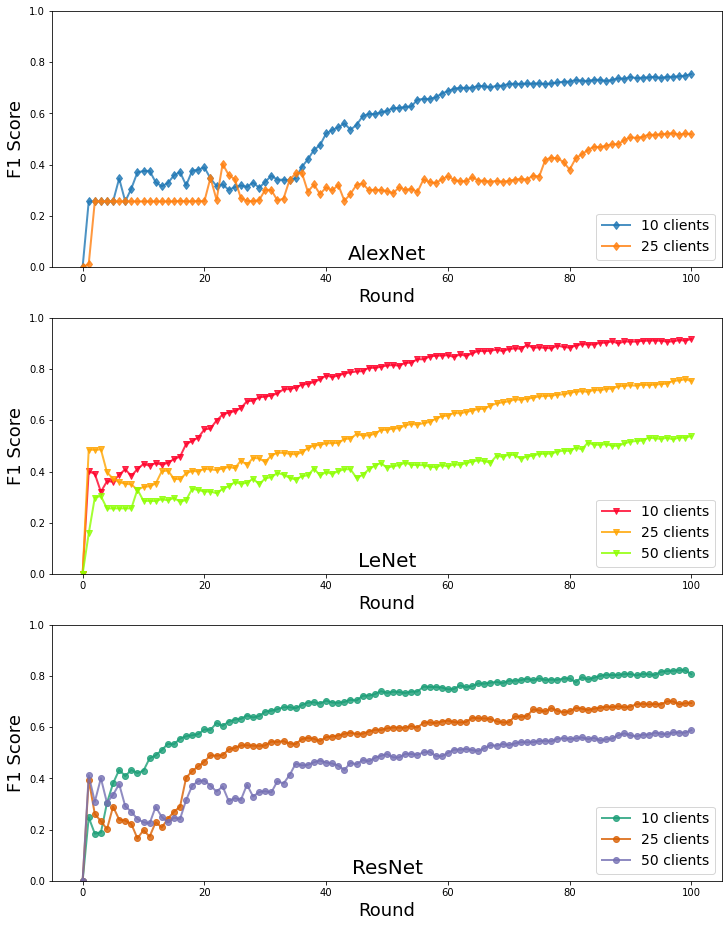}
    \caption{Effect of varying number of \(clients \in \{10,25,50\}\). \(c_{fraction}\) of 0.5, and \(40\%\) skewness on the less common labels were used here, however, since the number of clients is more than the number of unique labels, the distribution over the clients ends up more IID.} 
    \label{fig:n_clients}
\end{figure}

We vary the number of clients between 10, 25, and 50, with the client fraction set to 0.5 for all the runs. This effectively means approximately half of the data is used for training on each round. Since the client numbers are large, we have to use the less common labels to have an equal number of data in all the clients, thus effectively making the data IID in nature. From Fig \ref{fig:n_clients}, it is quite clear that increasing the number of clients impacts the convergence quite drastically. All three models converge faster and better for \(n=10\). Next, there is a drop in F1-score for \(n=25\), and a further drop for \(n=50\). In the case of AlexNet, given that it is a very big model, our hardware restrictions did not allow us to scale beyond 30 clients, and hence the experiment for \(n=50\) was not completed.

\begin{figure}[h]
    \subfigure[]{\includegraphics[width=0.9\linewidth]{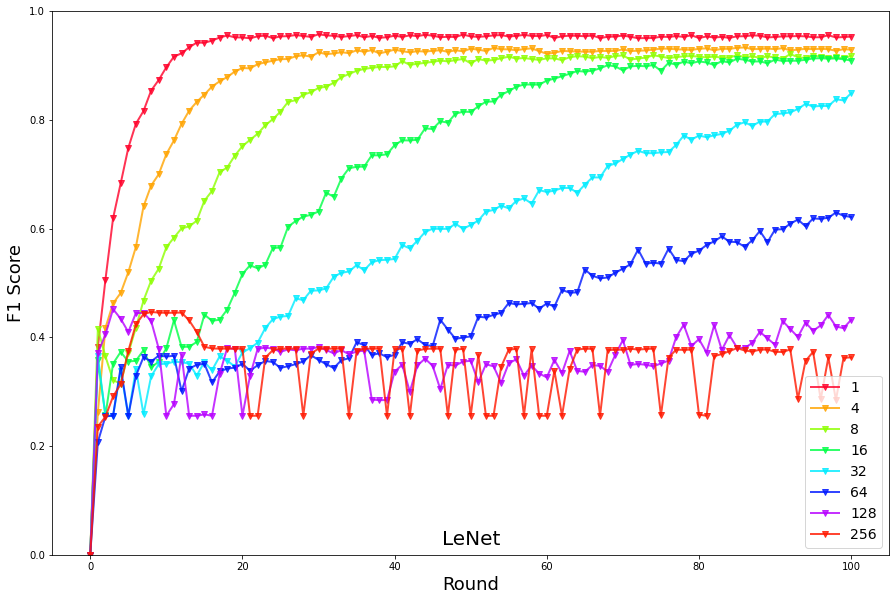}}
    \subfigure[]{\includegraphics[width=0.9\linewidth]{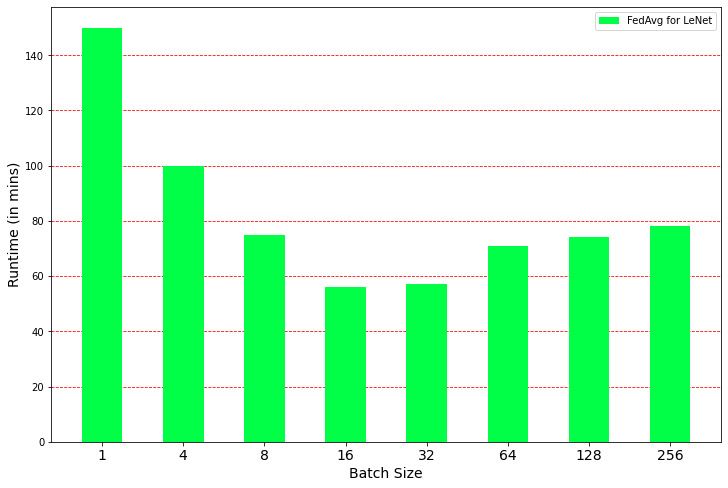}}
    \caption{(a) shows the effect of convergence for different batch sizes, with 4 clients and \(40\%\) skewness on common labels. (b) shows the runtime for 100 rounds for different batch sizes. The training runs are for LeNet.}
    \label{fig:bs_comparison}
\end{figure}

\subsubsection{Batch Size}
Batch size is varied between 1, 4, 8, 16, 32, 64, 128, and 256. We use 4 clients with  \(c_{fraction} = 0.75\) for these experiments. Increasing the batch size affects the convergence of the deep learning model conversely, as seen from Fig. \ref{fig:bs_comparison}(a). For a batch size of 1, the model converges the quickest and to the highest score. With an increase in the batch size, the model consistently takes longer to converge and converges to lower F1-scores. The biggest drop in F1-score is between 16 and 32, where the F1 score drops by around 22\%, and for larger batch sizes (64, 128, 256), the model fails to converge to any meaningful results. 
\\
While it is quite evident that batch size 1 performs the best, when looking into the bar graph presented in Fig. \ref{fig:bs_comparison}(b), we see that the run times are very different for different batch sizes. The runtimes presented are for 100 rounds. A batch size of 1 takes around 2.5 times longer than for a batch size of 16, where the drop in F1-score between them is ~3.5\%. So, depending on the application and the efficiency and hardware required, it could be more suitable to use larger batch sizes for a drop of a few accuracy points. Further increase in batch size leads to a slight increase in run times, and this could be due to the overhead costs due to memory restrictions. While the empirical results show that batch size 16 might be an ideal balance between run time and F1 score, this is only for LeNet model, and the optimum batch size number heavily depends on the deep learning model used. These results could vary for AleXNet and ResNet.

\subsubsection{Data Skewness between Clients}
\begin{figure}[h]
\centering
    \includegraphics[width=\linewidth]{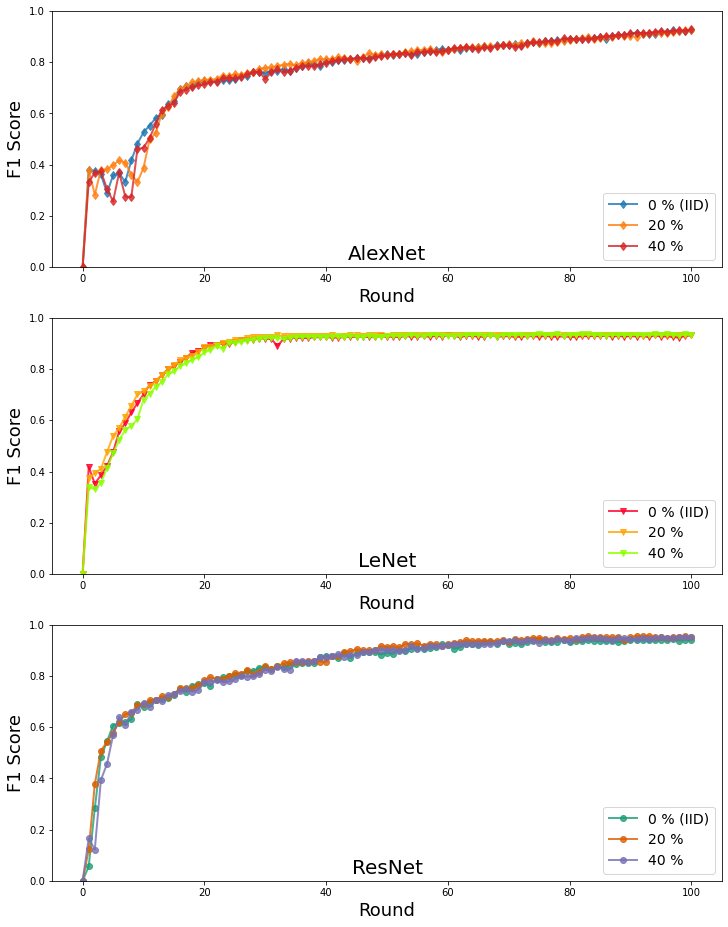}
    \caption{Effect of varying data \(skewness \in \{0,20,40\}\% \) on common labels on LeNet. 4 clients with a \(c_{fraction}\) of 0.75 were used.}
    \label{fig:skewsize_common}
\end{figure}

Initially, we use common labels for splitting the data to generate a non-IID distribution. The initial baseline for these experiments are 0\% skewness, which represents the IID data distribution. Next we increase the skewness to 20\% and 40\%. The number of clients used is 4, and for our dataset, there was no apparent difference in the convergence for these degrees of skewness. As seen in Fig. \ref{fig:skewsize_common}, on all the 3 Deep Learning models, the model convergence for different skewnesses is very similar and converges also to similar final F1-score values. This indicates that FedAvg is able to handle low levels of non-IIDness in the data quite well. Next, we further increase the skewness to higher values of 60\% and 80\%. For this, we will have to use a higher number of clients and the less common labels as explained in section \ref{section:exps}(iii).

\begin{figure}[h]
\centering
    
    \subfigure[]{\includegraphics[width=\linewidth]{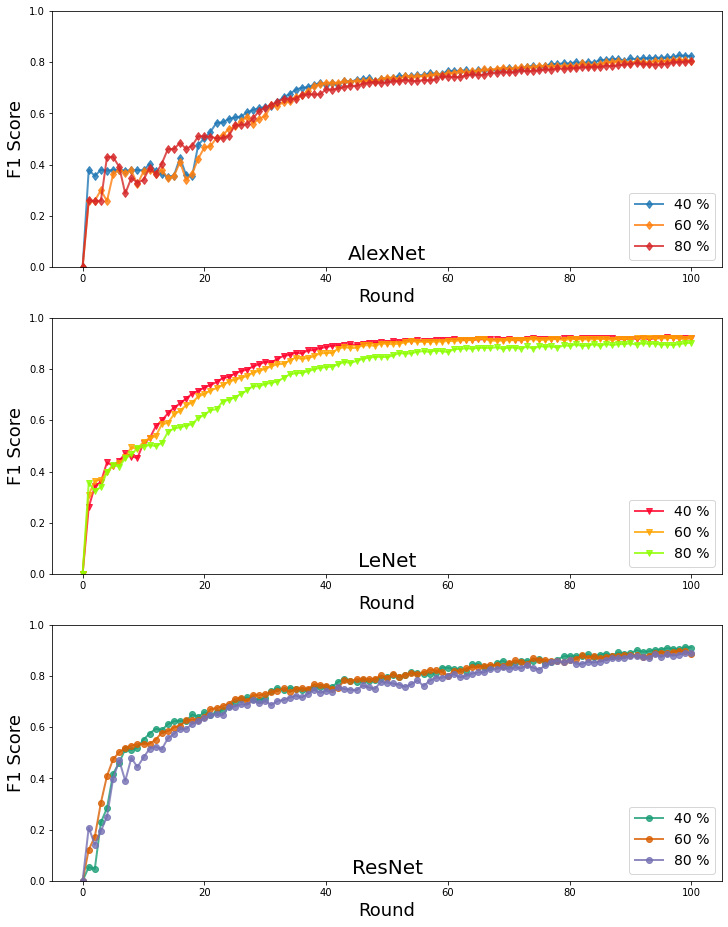}}
    \subfigure[]{\includegraphics[width=\linewidth]{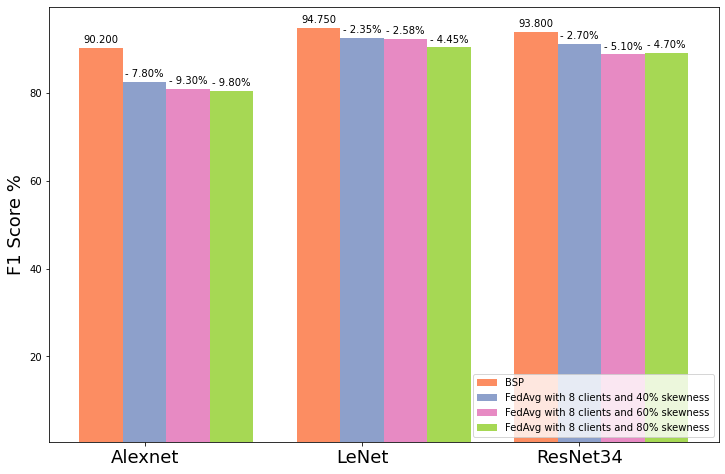}}
    \caption{(a) shows the effect of varying data  \(skewness \in \{40,60,80\}\% \) on less common labels on LeNet. 8 clients with a \(c_{fraction}\) of 0.75 were used. (b) shows the difference between the max F1 scores achieved by BSP and FedAvg.}
    \label{fig:skewsize}
\end{figure}

With 8 clients, we use skewness of 40\%, 60\%, and 80\%. These learning curves are shown in  Fig \ref{fig:skewsize}(a). Again, we see very similar learning curves, but there is a slight drop in the learning curves' convergence time for skewness of 80\% in all the 3 models. This is especially seen clearly in the LeNet learning curves where convergence takes longer than the lower skewness case. To showcase the difference better, we present the maximal F1-scores of the three different skewness degrees in Fig. \ref{fig:skewsize}(b) for the three Deep Learning models compared to BSP for the same skewness settings. We can see that there is a drop of F1-score with an increase in skewness, albeit slightly. Overall, for the dataset we have used, we can summarise that the skewness of the data does not impact the learning of the Deep Learning models using FedAvg. These results might vary when using a larger dataset.

\begin{figure}[h]
\centering
    \subfigure[]{\includegraphics[width=\linewidth]{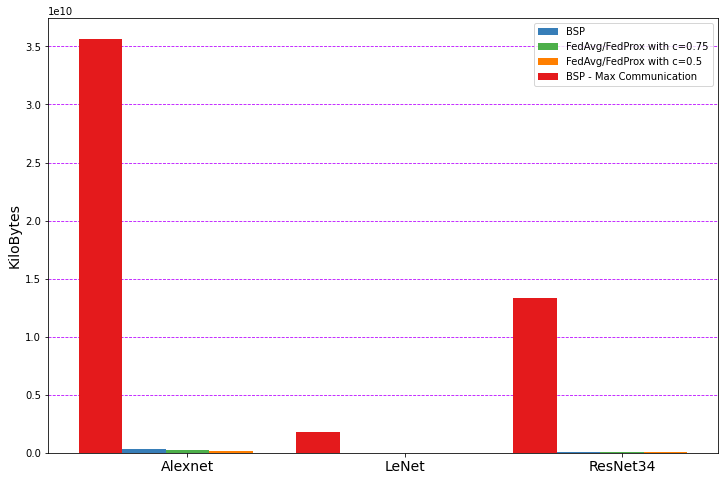}}
    \subfigure[]{\includegraphics[width=\linewidth]{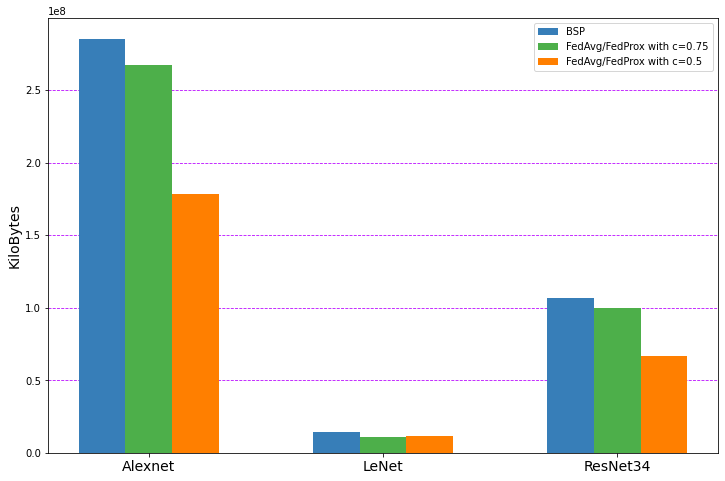}}
    \caption{Both graphs show the training communication costs in KiloBytes on our dataset: (a) shows the cost in comparison to a BSP-max (in red), where the communication between the server and the client is happening after each training batch, which is very costly compared to our BSP (in Blue) or FedAvg/FedProx. (b) shows the communication difference between our BSP (where communication happens after training on all the batches of a client) and FedAvg/FedProx with \(C_{fraction} \in \{0.5,0.75\}\)}
    \label{fig:commcost_exps}
\end{figure}

\subsection{Communication Cost Comparison}
\label{section:commcost}
In this section, we present the communication costs associated with the different federated algorithms and try to evaluate the optimal trade-off between accuracy and communication cost among the different setups. 

In Fig \ref{fig:commcost_exps}(a), we see a comparison of BSP at maximum communication mode (i.e., when a model is moved from one client to another after each batch of data) with other methods of federated learning. The communication costs are so high that the other methods are almost unnoticeable on the bar graph. Next, we compare BSP with a better efficiency technique, where the model is moved after training on the entire partition of a client to the next client. \textit{Federated Averaging} and \textit{Federated Proximal} methods have the same communication cost, and hence they are shown using the same bars. The client fraction for these two federated algorithms is varied. Lower client fraction means that the model is moved to a smaller number of clients, and this translates to a linear reduction in communication cost with lowering the client fraction. BSP still has the highest communication cost even after the optimization, but the difference now is much smaller. The problem with BSP, however, is that the model has to be trained sequentially on each client, and this means the runtime on BSP is again high compared to FedAvg, where the training can happen on $n$ different clients at once. The communication costs also depend heavily on the deep learning model used. In the case of LeNet, the model communication cost is quite low since it's a smaller network, but ResNet and AlexNet require more communication to be trained well. Depending on the size and nature of the dataset, we can opt for different Deep Learning networks. For our use case, LeNet was able to classify the data well and is generally also the most optimal, with communication costs in mind.

\section{Conclusion}
We present the findings and results of applying federated learning for multi-label image classification on a remote sensing dataset. Federated learning has known advantages that become more relevant for remote sensing cases, and our experiments certainly show that federated learning can be a useful training solution for remote sensing even when the data on different clients is non-IID in nature.

We evaluate three different federated learning algorithms: Bulk Synchronous Parallel (BSP), Federated Averaging (FedAvg), and Federated Proximal (FedProx) using three Deep Convolutional networks \textit{LeNet}, \textit{AlexNet} and \textit{ResNet34}. BSP performs the best among the three federated algorithms, but given its high communication costs and/or runtimes for a practical use case, FedAvg and FedProx might be more suitable. Albeit a slight drop in F1-score, these algorithms achieve results quite efficiently and also provide a parameter called client fraction, which can be used to control the trade-off between communication cost and accuracy. For the UC Merced Land Use dataset LeNet performed the best in our experiments. We also discussed the effect of varying different hyperparameters on the overall model convergence and presented the best practices for the same.

\subsection{Future Work}
In the future, we would like to test out the experiments on a bigger dataset, as this would help to validate these results on a larger scale. We speculate that using larger datasets might also give different results when it comes to a high data skewness use case. 

We would also like to experiment on a more complex dataset, where the remote sensing images have more channels than the RGB images in UC Merced Land Use dataset. One dataset that could suffice both size and complexity requirement could be BigEarthNet, which contains more than 500 thousand 13-channels images\cite{DBLP:BigEarthNet}. 

We also plan to implement another federated learning approach that manipulates the gradients rather than the weights to handle client divergence. Deep Gradient Compression \cite{lin2020deep} is an ideal candidate for such a method. This will give us more insight into which federated algorithm works for which application.

\subsection{Acknowledgements}
This work was supported by our mentor Arne de Wall. We thank him for all his feedback and for his technical support and tips to achieve the results.

\nocite{non_iid_main_paper}
\nocite{federatedLearningGoogle}
\nocite{fedAvg}
\nocite{fedProx}
\bibliography{maincitation}

\begin{appendices}
\begin{table*}[ht]
\caption{Experimental Setup: Parameters}
\label{table:experiments}

\centering
\begin{tabular}{p{0.1\linewidth}p{0.1\linewidth}p{0.05\linewidth}p{0.05\linewidth}p{0.1\linewidth}p{0.05\linewidth}p{0.1\linewidth}p{0.05\linewidth}p{0.05\linewidth}}
\hline
DL Model & FL Algorithm & Epochs & Clients & Batch Size & C-Fraction & Skewness & Client Epochs & Small Skew \\
\hline
LeNet & Centralized & 100 & NA & 4 & NA & NA & NA & NA\\
ResNet & Centralized & 100 & NA & 4 & NA & NA & NA & NA\\
AlexNet & Centralized & 100 & NA & 4 & NA & NA & NA & NA\\
LeNet & BSP & 100 & 8 & 4 & 0.5 & 40 & 5 & TRUE\\
ResNet & BSP & 100 & 8 & 4 & 0.5 & 40 & 5 & TRUE\\
AlexNet & BSP & 100 & 8 & 4 & 0.5 & 40 & 5 & TRUE\\
\hline
LeNet & FedAvg & 100 & 8 & 4 & 0.5 & 40 & 5 & TRUE\\
LeNet & FedAvg & 100 & 8 & 4 & 0.75 & 40 & 5 & TRUE\\
LeNet & FedAvg & 100 & 8 & 4 & 1 & 40 & 5 & TRUE\\
ResNet & FedAvg & 100 & 8 & 4 & 0.5 & 40 & 5 & TRUE\\
ResNet & FedAvg & 100 & 8 & 4 & 0.75 & 40 & 5 & TRUE\\
ResNet & FedAvg & 100 & 8 & 4 & 1 & 40 & 5 & TRUE\\
AlexNet & FedAvg & 100 & 8 & 4 & 0.5 & 40 & 5 & TRUE\\
AlexNet & FedAvg & 100 & 8 & 4 & 0.75 & 40 & 5 & TRUE\\
AlexNet & FedAvg & 100 & 8 & 4 & 1 & 40 & 5 & TRUE\\
\hline
LeNet & FedAvg & 100 & 8 & 4 & 0.75 & 60 & 5 & TRUE\\
LeNet & FedAvg & 100 & 8 & 4 & 0.75 & 80 & 5 & TRUE\\
ResNet & FedAvg & 100 & 8 & 4 & 0.75 & 60 & 5 & TRUE\\
ResNet & FedAvg & 100 & 8 & 4 & 0.75 & 80 & 5 & TRUE\\
AlexNet & FedAvg & 100 & 8 & 4 & 0.75 & 60 & 5 & TRUE\\
AlexNet & FedAvg & 100 & 8 & 4 & 0.75 & 80 & 5 & TRUE\\
\hline
LeNet & BSP & 100 & 4 & 4 & 0.75 & 40 & 5 & FALSE\\
AlexNet & BSP & 100 & 4 & 4 & 0.75 & 40 & 5 & FALSE\\
ResNet & BSP & 100 & 4 & 4 & 0.75 & 40 & 5 & FALSE\\
LeNet & FedAvg & 100 & 4 & 4 & 0.75 & 40 & 5 & FALSE\\
AlexNet & FedAvg & 100 & 4 & 4 & 0.75 & 40 & 5 & FALSE\\
ResNet & FedAvg & 100 & 4 & 4 & 0.75 & 40 & 5 & FALSE\\
LeNet & FedAvg & 100 & 4 & 4 & 0.75 & 20 & 5 & FALSE\\
AlexNet & FedAvg & 100 & 4 & 4 & 0.75 & 20 & 5 & FALSE\\
ResNet & FedAvg & 100 & 4 & 4 & 0.75 & 20 & 5 & FALSE\\
LeNet & FedAvg & 100 & 4 & 4 & 0.75 & 0 & 5 & FALSE\\
AlexNet & FedAvg & 100 & 4 & 4 & 0.75 & 0 & 5 & FALSE\\
ResNet & FedAvg & 100 & 4 & 4 & 0.75 & 0 & 5 & FALSE\\
\hline
LeNet & FedAvg & 100 & 10 & 4 & 0.5 & 40 & 5 & TRUE\\
AlexNet & FedAvg & 100 & 10 & 4 & 0.5 & 40 & 5 & TRUE\\
ResNet & FedAvg & 100 & 10 & 4 & 0.5 & 40 & 5 & TRUE\\
LeNet & FedAvg & 100 & 25 & 4 & 0.5 & 40 & 5 & TRUE\\
AlexNet & FedAvg & 100 & 25 & 4 & 0.5 & 40 & 5 & TRUE\\
ResNet & FedAvg & 100 & 25 & 4 & 0.5 & 40 & 5 & TRUE\\
LeNet & FedAvg & 100 & 50 & 4 & 0.5 & 40 & 5 & TRUE\\
AlexNet & FedAvg & 100 & 50 & 4 & 0.5 & 40 & 5 & TRUE\\
ResNet & FedAvg & 100 & 50 & 4 & 0.5 & 40 & 5 & TRUE\\
\hline
LeNet & FedProx & 100 & 8 & 4 & 0.75 & 40 & 5 & TRUE\\
AlexNet & FedProx & 100 & 8 & 4 & 0.75 & 40 & 5 & TRUE\\
ResNet & FedProx & 100 & 8 & 4 & 0.75 & 40 & 5 & TRUE\\
LeNet & FedProx & 100 & 25 & 4 & 0.5 & 40 & 5 & TRUE\\
AlexNet & FedProx & 100 & 25 & 4 & 0.5 & 40 & 5 & TRUE\\
ResNet & FedProx & 100 & 25 & 4 & 0.5 & 40 & 5 & TRUE\\
LeNet & FedProx & 100 & 10 & 4 & 0.5 & 40 & 5 & TRUE\\
AlexNet & FedProx & 100 & 10 & 4 & 0.5 & 40 & 5 & TRUE\\
ResNet & FedProx & 100 & 10 & 4 & 0.5 & 40 & 5 & TRUE\\
\hline
LeNet & FedAvg & 100 & 4 & 1 & 0.75 & 40 & 5 & FALSE\\
LeNet & FedAvg & 100 & 4 & 4 & 0.75 & 40 & 5 & FALSE\\
LeNet & FedAvg & 100 & 4 & 8 & 0.75 & 40 & 5 & FALSE\\
LeNet & FedAvg & 100 & 4 & 16 & 0.75 & 40 & 5 & FALSE\\
LeNet & FedAvg & 100 & 4 & 32 & 0.75 & 40 & 5 & FALSE\\
LeNet & FedAvg & 100 & 4 & 64 & 0.75 & 40 & 5 & FALSE\\
LeNet & FedAvg & 100 & 4 & 128 & 0.75 & 40 & 5 & FALSE\\
LeNet & FedAvg & 100 & 4 & 256 & 0.75 & 40 & 5 & FALSE\\
\hline
\end{tabular}
\end{table*}
\end{appendices}

\end{document}